\pdfoutput=1

\documentclass[11pt]{article}

\usepackage{acl}

\usepackage[utf8]{inputenc} 
\usepackage[T1]{fontenc}    
\usepackage{hyperref}       
\usepackage{url}            
\usepackage{booktabs}       
\usepackage{amsfonts}       
\usepackage{nicefrac}       
\usepackage{microtype}      
\usepackage{lipsum}
\usepackage{amsmath}
\usepackage{graphicx}
\graphicspath{{media/}}     
\usepackage{xcolor}
\usepackage{subcaption}
\usepackage{algorithm}
\usepackage{algpseudocode}
\usepackage{tcolorbox}
\usepackage{enumitem}
\usepackage{makecell}
\usepackage{float}

\usepackage{natbib}
\tcbuselibrary{breakable}

\title{An LLM-Based Approach for Insight Generation in Data Analysis
}

\author{
  Alberto Sánchez Pérez \\ 
  Aily Labs, EURECOM \\
  \fontsize{9.5pt}{9.5pt}\texttt{alberto.sanchez@ailylabs.com} \\
  \fontsize{9.5pt}{9.5pt}\texttt{sanchez@eurecom.fr} \\
  \And
  Alaa Boukhary \\ 
  Aily Labs \\
  \fontsize{9.5pt}{9.5pt}\texttt{alaa.boukhary@ailylabs.com} \\
  \AND
  Paolo Papotti \\
  EURECOM \\
  \fontsize{9.5pt}{9.5pt}\texttt{papotti@eurecom.fr} \\
  \And
  Luis Castejón Lozano \\
  Aily Labs \\
  \fontsize{9.5pt}{9.5pt}\texttt{luis.castejon@ailylabs.com} \\
  \And
  Adam Elwood \\ 
  Aily Labs \\
  \fontsize{9.5pt}{9.5pt}\texttt{adam.elwood@ailylabs.com} \\
}

\begin{document}
\maketitle
\begin{abstract}
Generating insightful and actionable information from databases is critical in data analysis. This paper introduces a novel approach using Large Language Models (LLMs) to automatically generate textual insights. Given a multi-table database as input, our method leverages LLMs to produce concise, text-based insights that reflect interesting patterns in the tables. Our framework includes a Hypothesis Generator to formulate domain-relevant questions, a Query Agent to answer such questions by generating SQL queries against a database, and a Summarization module to verbalize the insights. The insights are evaluated for both correctness and subjective insightfulness using a hybrid model of human judgment and automated metrics. Experimental results on public and enterprise databases demonstrate that our approach generates more insightful insights than other approaches while maintaining correctness.
\end{abstract}

\section{Introduction}

\begin{figure*}[h]
    \centering
        \includegraphics[trim=6.5mm 0mm 6.5mm 0mm, clip, width=0.8\textwidth]{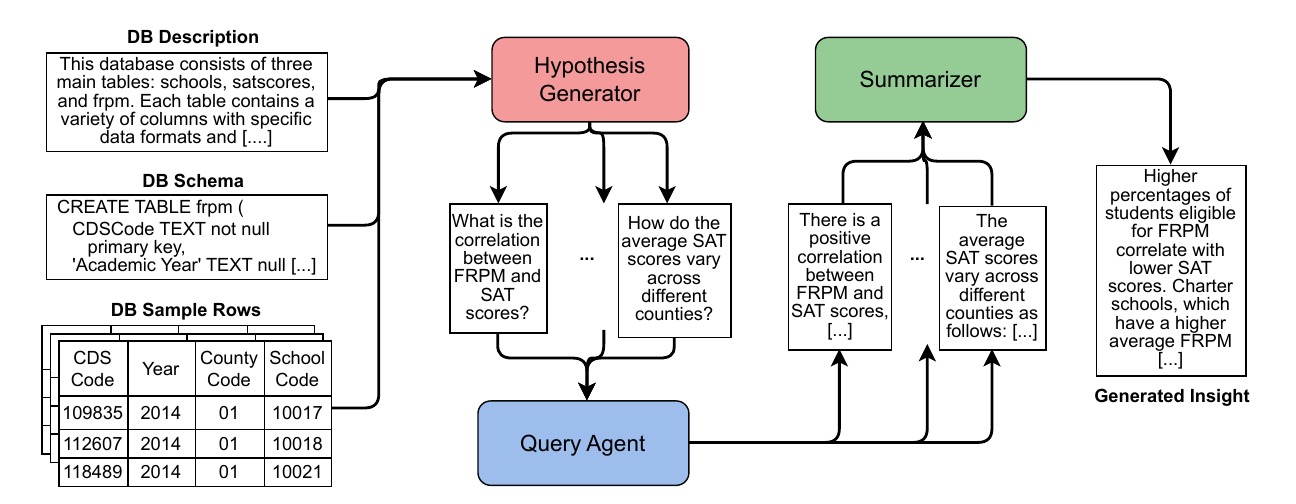}
        \caption{Overview of our approach: the Hypothesis Generator generates interesting, high-level questions, the Query Agent answers them using SQL scripts, and the Summarization module aggregates the results into an insight.}
        \label{fig:diagram}
\end{figure*}

In an era of data-driven decision-making, the ability to extract 
insights from complex databases has become increasingly important~\citep{veit_jahns_2013, stefan_steiner_2022}. These insights provide meaningful and actionable information derived from raw data. Insights are often derived through statistical, analytical, and computational methods. For example, given the database in Figure~\ref{fig:diagram}, 
a valuable insight would be \textit{``Higher percentages of students eligible for free or reduced-price meals (FRPM) correlate with lower average SAT scores in reading, math, and writing.''}

Insights are essential to make informed decisions. This is crucial in business contexts, 
healthcare, or scientific research, 
where better insights lead to more effective outcomes and advancements~\citep{attia_khan_2024, oluwatosin_abdulazeez__2024, andrew_ifesinachi_daraojimba__2024}.

Generating insights traditionally requires significant manual effort. Analysts must meticulously preprocess, explore, and refine the data. It also requires specialized knowledge of the domain to analyze and interpret the data, further complicating the task. This makes the process both time-consuming and resource-intensive~\citep{deepali_arora__2015, bean2022data}.

There have been multiple proposals for automating data analysis tasks~\citep{ma2023demonstration, ding2019quickinsights, openai2024data, langchain2024}. However, their outputs are not as insightful as human-crafted insights. They often only operate on single tables and require pre-cleaned data or user-defined goals, which limit their applicability to real-world use cases.

Our approach introduces a novel framework leveraging Large Language Models (LLMs) 
to automatically generate insights. However, the process of extracting valuable insights from raw data with LLMs is non-trivial. First, LLMs have limited context size and cannot be fed with entire databases~\citep{pawar2024whatwhycontextlength}. Second, LLMs still struggle in processing structured data~\citep{10.1145/3654979}. Because of these issues, we adopt an agentic approach, where external tools (SQL scripts, in our case) are used by the LLM to handle structured data. While this is a popular solution, existing methods mostly produce ``shallow'' insights, such as the identification of simple trends or outliers~\citep{openai2024data, langchain2024}. As depicted in  Figure~\ref{fig:diagram}, our idea is to obtain richer insights by first creating high-level questions over the database. Those questions are more complex than the insights that can be directly generated by the LLM. We exploit the fact that LLMs are able to split a question down into simpler ones that can be turned into SQL scripts and validated over the database. 
Finally, the query results are aggregated into a textual insight. 

The generated insights are evaluated using a hybrid approach that combines human judgment with automated metrics, ensuring a rigorous assessment of both correctness and insightfulness. Experiments on public and real company databases show that our solution clearly outperforms existing alternatives in terms of insightfulness of the output texts. 


\section{Preliminaries}
We first introduce Text-to-SQL as it plays a key role in our solution. We then define insights and the criteria for evaluating them.

\subsection{Text-to-SQL}
A database $D$ is defined as a set of tables $T_i$, each table is a set of tuples $T_i = {(t_{j1}, t_{j2}, \ldots, t_{jn}) \mid j \in 1 \ldots m}$ with $j$ representing the columns of the table. A query $q$ is a mapping between a list of input tables and an output table, such that $q: T_0 \times \ldots \times T_k \to T_O$, where $q$ is a sequence of operations, denoted as $Q = f_n \circ \ldots \circ f_0$. Each $f_i$ consists of relational operations (e.g., projection, selection, join), set operations (e.g., union, intersection), and set functions (e.g., sum, average). The result of applying query $q$ on database $D$ is denoted as $q(D) = R$, where $R$ is the resulting table.

The objective of Text-to-SQL can be defined as:
$\min \textit{dist}(q_{gen}(D), q_{GT}(D))$, where $q_{gen}$ is the query generated by the model, $q_{GT}$ is the ground truth query, which answers the question, and $\textit{dist}$ is a distance function between 
the resulting tables. 

\subsection{Insights}

We define an insight $I$ 
as a short text derived from a database $D$.
We set that it should not be longer than 3 sentences based on experiments with our product, which indicate that longer text is less effective in maintaining reader interest. 

Insights are derived using information from a subset or derivation of the database $f(D)$. For example, $f(D)$ can be the result of one or more SQL queries over $D$ or a transformation of $D$ through code.
In addition to $f(D)$, the input includes textual information $\textit{info}$: $\{f(D)\} \cup \textit{info} \to I$, with \textit{info} = $\{D_\text{info}, D_\text{schema}\}$, i.e.,  $D_\text{schema}$ contains the schema and a sample of rows of each table of $D$ (DB Schema and DB Sample Rows in Figure~\ref{fig:diagram}); $D_\text{info}$ is a textual high-level description of (i) the database, (ii) each table, (iii) the columns in each table (DB Description in Figure~\ref{fig:diagram}). If not available in the database catalog, This $D_\text{info}$ is LLM-generated with a prompt (Appendix~\ref{sec:dbdesc_prompt}), following the "verbose prompt" strategy~\citep{sun2024sqlpalmimprovedlargelanguage}.

\subsubsection{Insightfulness}
\label{sec:insightinsightfulness}

Insightfulness is a subjective metric of insights based on human judgment. It is hard to quantify objectively. This is due to insightfulness being case, domain, and user dependent. The metric is based on a set of implicit metrics $\{M_0(I, U),\ldots ,M_n(I, U)\}$ (e.g., actionability, relevance, novelty, \ldots) driven by expertise and KPIs. These metrics cannot be explicitly calculated given only $D$ as input and must be estimated by the user $U$, weighted by their preferences $\{w_{M_0}, \ldots, w_{M_n}\}$.
We then define insightfulness of an insight to a user as:
\[ \textit{insightfulness}(I, U) = \frac{\sum_{i=0}^n w_{M_i} \cdot M_i(I, U)}{\sum_{i=0}^n w_{M_i}} \]

\subsubsection{Correctness}
\label{sec:insightcorrectness}
Insights can be composed of one or more claims $C_i$. Claims are factual statements that have a truth value $TV$ that can be proven true or false, thus $TV(C_i) \in \{0,1\}$. We then define the correctness of I as the mean of the truth value of each of its claims:
\[ \textit{correctness}(I) = \frac{1}{n} \sum_{i=0}^n TV(C_i) \ \]

\section{Problem Definition}

The main objective of this work is the generation of short texts that are both \textit{insightful}
and \textit{correct}
(Sections~\ref{sec:insightcorrectness} and~\ref{sec:insightinsightfulness}).

Depending on the context, insightfulness and correctness might have different importance. We define the weighting factor $\alpha$ in the resulting function.
The objective is to maximize the weighted harmonic mean:
\[ O = \max \frac{1}{\frac{\alpha}{\textit{insightfulness}} + \frac{1-\alpha}{\textit{correctness}}} \]

By default, we assign insightfulness and correctness the same importance ($\alpha =0.5$).


\section{Proposed Architecture}
\label{sec:architecture}

\begin{figure*}[h]
    \centering
        \includegraphics[trim=6.5mm 0mm 6.5mm 0mm, clip, width=\textwidth]{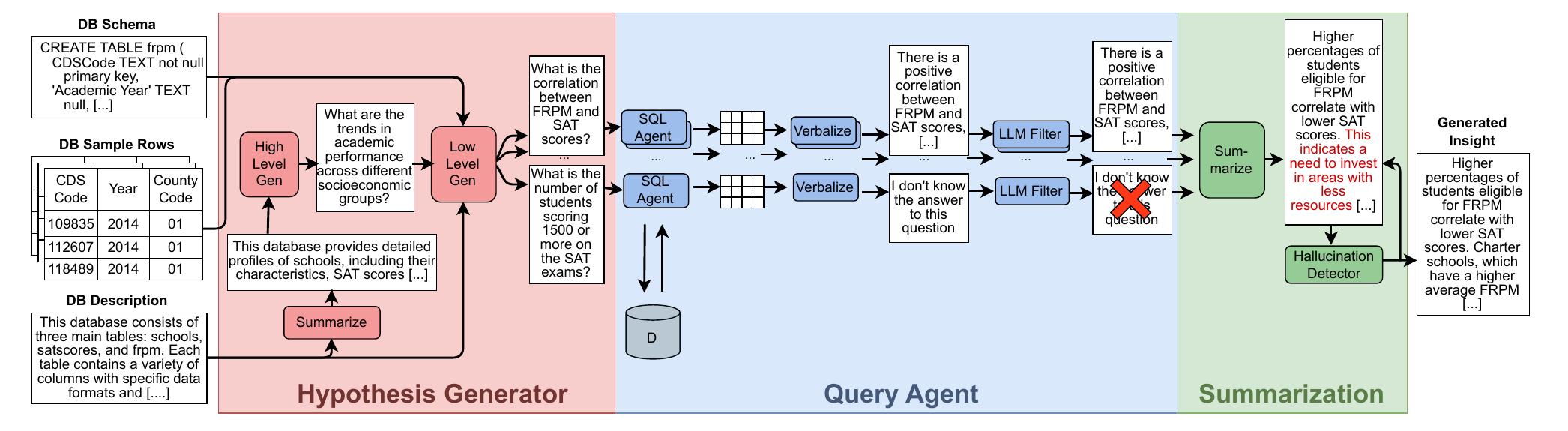}
        \caption{Proposed Architecture: The High-Level Generator generates questions using a short description
        , the Low-Level Generator splits each question into subquestions that are easier to answer by giving all the database details. The Query Agent uses SQL queries to answer those questions and validates them with LLM evaluation. Finally, the Summarizer aggregates the answers into a short insight and iteratively removes hallucinations to generate the final result.}
        \label{fig:diagram2}
\end{figure*}

We generate insights with a 3-step architecture, as in Figure~\ref{fig:diagram2}, that (1) generates high-level questions and splits each of them into simpler subquestions that are easier to answer, (2) then answers and validates them using SQL to (3) finally 
post-process them into the final insight.

\subsection{Hypothesis Generator}
The Hypothesis Generator uses a multi-step approach to generate questions that are both interesting and easy to answer for the given database. 
It is composed of two elements.
A high-level generator $\text{HL-G}$ that uses a prompt (Appendix~\ref{sec:hli_prompt}) to generate overarching questions $h_i$. The high-level generator uses a short description of the database $(\textit{short}(D_\text{info}))$  that we shorten to prevent constraining the model's response with database details, enabling more exploratory questions within the domain:
\[\text{HL-G}(\textit{short}(D_\text{info}))\to h_0,\ldots, h_n\]
    
A low-level generator $\text{LL-G}$ splits the high-level question $h_i$ into subquestions $s_{ij}$. By using specific prompting (Appendix~\ref{sec:hli_prompt}) and the full description and schema of the database as input, we constrain the model to formulate more precise and concrete questions, which are easier to answer in the next step:
\[\text{LL-G}(h_i, D_\textit{info}, D_\textit{schema})\to s_{i0},\ldots,s_{im}\]

\subsection{Query Agent}
Our work employs SQL queries instead of pandas to overcome the speed, storage, and scalability limitations of dataframes.
The Query Agent \textit{QAgent} generates a SQL query $q_{ij}$ and resulting tables $R_{ij}$ ($q_{ij}(D) = R_{ij}$) to answer the subquestions using the database information and schema:
\[ \textit{QAgent}(s_{ij}, D_\textit{info}, D_\textit{schema}) \to \ q_{ij}\]

The objective of the \textit{QAgent} is to minimize $\textit{dist}(q_{ij}(D), q^*_{ij}(D))$, where $q^*_{ij}$ is the ground truth query and $\textit{dist}$ is a distance function between tables based on the cell metrics in~\citep{papicchio2023qatch}. More concretely, we define $\textit{dist}$ as the harmonic mean of the metrics \textit{cell-precision} and \textit{cell-recall}, \textit{dist} = $\frac{2 \cdot \textit{cell-precision} \cdot \textit{cell-recall}}{\textit{cell-precision} + \textit{cell-recall}}$.

Queries are then verbalized $\textit{verb}(R_{ij})$ in natural language with a prompt (Appendix~\ref{sec:sqlagent_prompt}) to answer $s_{ij}$. They are then validated using LLM evaluation functions that use an LLM score, $\textit{score}_a$ of answer relevance and answerability~\citep{lin2023llmevalunifiedmultidimensionalautomatic}, to remove questions with a score under a threshold $\tau_a$\footnote{The experimentally determined value of $\tau_a$ is 0.7 for both answerability and relevance, effectively filtering most low-quality answers while retaining a sufficient number of relevant ones.}.

\subsection{Summarization}
The Summarization step summarizes ($\textit{summ}$) the verbalized answers to generate the Insight:
\[\textit{summ}(\textit{verb}(R_{i0}),\ldots,\textit{verb}(R_{im})) \to I \ ; \]

\begin{algorithm}[H]
\caption{Iterative Reflection Based on Verb Relations} \label{alg:filter}
\small
\begin{algorithmic}[1]
\State $i \gets 0$
\While{$\textit{score}_h(I, \{\textit{verb}(R_{i0}), \ldots, \textit{verb}(R_{im})\}) \geq \tau_a$ \textbf{and} $i < \textit{maxit}$}
    \State $\textit{reflect}(I, \{\textit{verb}(R_{i0}), \ldots, \textit{verb}(R_{im})\}) \to I$
    \State $i \gets i + 1$
\EndWhile
\State \textbf{return} $I$
\end{algorithmic}
\normalsize
\end{algorithm}

Each generated insight is post-processed to filter out possible hallucinations, as detailed in Algorithm~\ref{alg:filter}.
An evaluation function $\textit{score}_h$ is used to measure an LLM hallucination score by splitting the insight into different claims and using a LLM to generate a score based on the contradictions between them and the answers $\textit{verb}(R_{ij})$~\citep{liu2023gevalnlgevaluationusing}. The summary uses reflexion~\citep{shinn2023reflexionlanguageagentsverbal} to iteratively correct the summary until the LLM hallucination score is under a threshold $\tau_h$\footnote{The value of $\tau_h$ has been experimentally determined at 0.9 to ensure rigorous hallucination removal.} or exceeds an iteration limit $\textit{maxit}$.


\section{Evaluation}

Our proposed framework was evaluated on both \textit{insightfulness} (Section~\ref{sec:insightinsightfulness}) and \textit{correctness} (Section~\ref{sec:insightcorrectness}) using a combination of human and LLM evaluation.

\subsection{Experimental Setup}

\noindent{\bf Datasets.}
We generated a \textit{private dataset} from the company's internal data. This is used to evaluate  
the approach in real-world, proprietary contexts. It also allows for in-depth evaluation using domain experts from the company.
    \begin{itemize}
        \item \textbf{private\_sales:} internal sales and sales forecasting data; it has 3 tables with 16-49 columns (median of 17).
    \end{itemize}
    
We also employ a set of public datasets to ensure reproducibility and for a broader evaluation in a range of domains.
These datasets a sampled from the BIRD benchmark~\citep{li2023llm}, which is composed from real data sources 
spanning 37 domains~\citep{volvovsky2024dfinsqlintegratingfocusedschema}. 

We focus on a subset rather than the entire database benchmark to manage costs, as the \textit{correctness} evaluation requires specialized human judgment. In the sampling process, we ensured thematic and structural variety across the databases by including:
    \begin{itemize}
        \item \textbf{california\_schools:} education data in the state of California; it has 3 tables with 11-49 columns (median of 29).
        \item \textbf{codebase\_community:} data from posts in the social media platform Reddit; it has 8 tables with 4-21 columns (median of 6).
        \item \textbf{debit\_card\_specializing:} card transactions data in the fuel industry; it has 5 tables with 2-9 columns (median of 3).
        \item \textbf{european\_football\_2:} football data with teams, players, and matches; it has 7 tables with 2-115 columns (median of 7).
        \item \textbf{student\_club:} data from a student club with financial data, attendance metrics and events; it has 8 tables with 2-9 columns (median of 6).
    \end{itemize}

\noindent{\bf LLMs.} All methods use the same base model, GPT4o~\citep{openai_gpt4o_2024},
to have a fair comparison between them. 
We use GPT4o because of its effectiveness in generating structured outputs with Function Calling~\citep{openai2024}. The average cost of generating an insight with our method amounts to 63 cents, 
which significantly reduces the cost from manual insight generation of \$140, i.e., the average estimated cost at our own company for a human-crafted insight (see Appendix~\ref{sec:cost} for more details). Also, potential decreases in LLM processing costs in the future will further reduce the cost of the system.

For the Text-to-SQL model, we 
adopt the LangGraph Sql Agent~\citep{langchain_sql_agent}. Following best practices to generate the queries~\citep{rajkumar2022evaluating}, the schema and 3 sample rows are given to the model.

\vspace{1ex}
\noindent{\bf Baselines and Models.}
\label{sec:baselines}
We compare our solution against a set of baselines. 
    \begin{itemize}
        \item \textbf{GPT-DA:} Insights are generated using the ChatGPT data analysis functionality with the prompt detailed in Appendix~\ref{sec:baseline_prompt}. It takes as input a CSV file and uses pandas code generation to output insights~\citep{openai2024data}.
        \item \textbf{Quick:} Insights are generated using QuickInsights from MS PowerBI~\citep{powerbi}. Results are obtained from pre-joined if possible. If not, from random individual tables~\citep{ding2019quickinsights}.
        \item \textbf{Serial:} 
        Insights are generated by serializing a subset of the database into an HTML table due to context windows limitations in the LLMs~\citep{pawar2024whatwhycontextlength}, which is then fed into the LLM prompt detailed in Appendix~\ref{sec:baseline_prompt}.
    \end{itemize}

We also evaluate variations of our architecture described in Section~\ref{sec:architecture} as an ablation study.
    \begin{itemize}
        \item \textbf{High-Level Insights ({\sc HLI}):} Insights are generated with high-level and low-level questions as described in Section~\ref{sec:architecture}.
        \item \textbf{High-Level Insights without Summarization of the description ({\sc HLI-wS}):} Same architecture as the {\sc HLI} method, but the High-Level Generator uses the full database description as an input, instead of a short summary.
        \item \textbf{High-Level Insights without High-Level ({\sc HLI-wH}}): Insights are generated directly with low-level questions, without high-level questions. 
    \end{itemize}
As detailed in Appendix~\ref{sec:insight_number}, the number of generated insights varies across different methods, influenced by characteristics of each method, like filtering or data preprocessing. Some methods like {\sc HLI}, {\sc HLI-wH} or {\sc HLI-wS} have stricter filtering, leading to fewer total insights (30-40 in total), while existing methods have a slightly higher number of insights (40-50).

\subsection{Insightfulness Evaluation}
Due to the difficulty of measuring $\textit{insightfulness}(I,U)$, we use a relative measure to better estimate it. We define a ranker $R(U,I_A,I_B)$ that classifies the insight with the most insightfulness such that:
\[
R(U, I_A, I_B) =
\begin{cases}
    I_A & \text{if } \textit{insightfulness}(I_A,U) > \\
        & \quad \textit{insightfulness}(I_B,U) \\
    I_B & \text{otherwise}
\end{cases}
\]

We then define the $\textit{insightfulness}$ based on an Elo rating system as follows:
\begin{align}
    & \textit{insightfulness}(I, U) = \nonumber \\
    & \text{Elo}(I \ | \ R_0(U, I_{A0}, I_{B0}),\ldots,R_k(U, I_{Ak}, I_{Bk})) \ ; \nonumber
\end{align}

Where 
the Elo rating of $I$ is based on a list of pair-wise insight comparisons from the ranker $R_0(U, I_{A0}, I_{B0}),\ldots,R_k(U, I_{Ak}, I_{Bk})$.
For example, the insight ``Higher percentages of students eligible for free or reduced-price meals (FRPM) correlate with lower average SAT scores in reading, math, and writing.'' would get a better score than ``Los Angeles County has the highest number of students eligible for free meals, totaling 898,610 students.''.

All compared insights are sampled from a uniform distribution:
\[ I_{A0},I_{B0},\ldots,I_{Ak},I_{Bk}\sim\text{Uniform}(\{I_0,\ldots,I_n\}) \]

\begin{figure}[h]
    \includegraphics[width=0.5\textwidth]{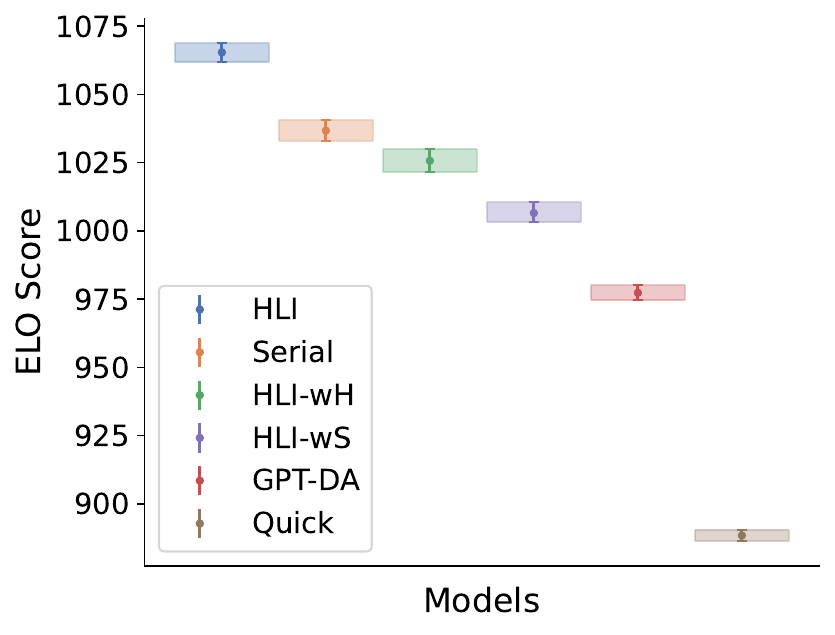}
    \caption{\textbf{Human} Elo Bootstrap Confidence Intervals (95\%) in the database private\_sales. Each bootstrap sample has a different order of comparisons.}
    \label{fig:insightfulness_human_ci}
\end{figure}

\begin{figure}[h]
    \includegraphics[width=0.5\textwidth]{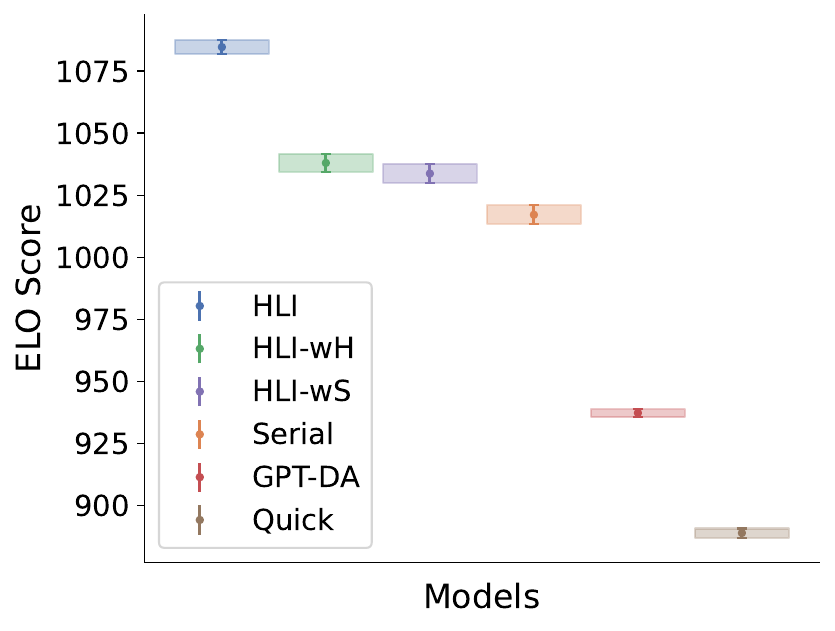}
    \caption{\textbf{LLM} Elo Bootstrap Confidence Intervals (95\%) in the database private\_sales. Each bootstrap sample has a different order of comparisons.}
    \label{fig:insightfulness_llm_ci}
\end{figure}

\begin{figure*}[h]
    \includegraphics[width=\textwidth]{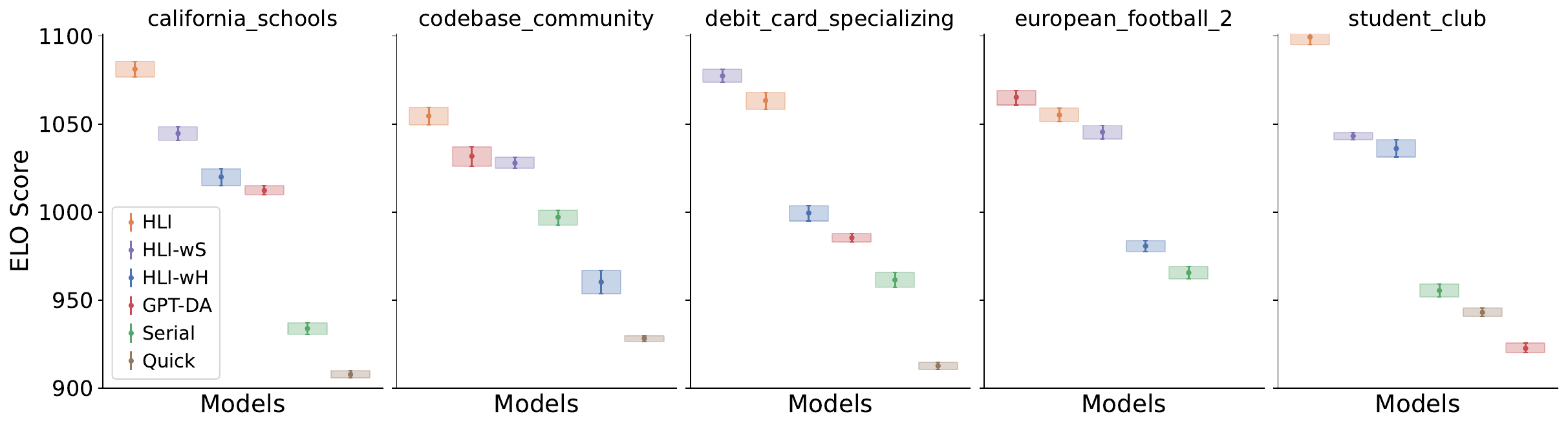}
    \caption{\textbf{LLM} Bootstrap Confidence Intervals (95\%) in public BIRD databases. Each bootstrap sample has a different order of comparisons.}
    \label{fig:insightfulness_llm_ci_publicdbs}
\end{figure*}

The Elo score has proven useful in translating subjective pairwise comparisons into numeric scores to rank LLM performance~\citep{chiang2024chatbot, askell2021general, boubdir2023elo}.
After each insight comparison, the ranking $R$ for each of the models $A$ is updated in relation to the other model $B$ based on the result of the comparison ($S_A = 1 \text{ if } R(U, I_A, I_B) = I_A \ ; S_A=0 \text{ else}$), using the formula
\[ R'_A = R_A + K \cdot (S_A - E_A) \]
where $E_A$ is the expected score of the model according to the formula:
\[ E_A = \frac{1}{1 + 10^{(R_B - R_A) / 400}} \]
In our implementation, we assigned each model an initial rating of 1000 and a k-factor $K$ of 4 to stabilize the Elo ratings and avoid biases towards recent games~\citep{GoogleColab2024}\footnote{Intuitively, the rating differential between two LLMs acts as a predictor for the comparison outcome. For two LLMs with identical ratings, they are expected to achieve an equal number of victories. If an LLM's rating exceeds its opponent's by 100 points, the first LLM is predicted to win 64\% of the time. If the rating differential is 200 points, the expected winning percentage for the higher-rated LLM rises to 76\%.}. However, Figure~\ref{fig:insightfulness_human_ci}, \ref{fig:insightfulness_llm_ci} and \ref{fig:insightfulness_llm_ci_publicdbs} represent 95\% confidence intervals with bootstrapping using different order of the comparisons. In those cases, we use a k-factor of 8 for better visualization to get slightly bigger intervals, as it has minimal effects on the final ordering after 100 comparisons,
 more details reported in Appendix~\ref{section:kfactor}.

\noindent{\bf Human Evaluation.}
For the human evaluation, we tasked a domain expert with performing 100 comparisons between insights generated by different methods on the private dataset.
The human evaluation for correctness and insightfulness metrics was conducted separately, with a domain expert dedicating approximately one hour to evaluate insightfulness.
As shown in Figure~\ref{fig:insightfulness_human_ci}, our method, {\sc HLI}, achieves the best results in terms of insightfulness. This comparison also works as an ablation test, showing the importance of the high-level and low-level question split ({\sc HLI-wH}), as the split allows for deeper high-level questions that are answered through different perspectives with each subquestion.
We also show that giving less information to the high-level generator ({\sc HLI-wS}) allows for better \textit{insightfulness}, as the high-level questions are less constrained due to not knowing the details of the database, allowing for more exploratory questions.

\vspace{1ex}
\noindent{\bf LLM Evaluation.}
To scale the evaluation of different insight generation models over a range of domains, we propose a new LLM-based evaluation model (the prompt for this task is reported in Appendix~\ref{sec:llm_eval_prompt}).

\begin{table}[h]
\centering
\setlength{\tabcolsep}{4pt}
\resizebox{0.48\textwidth}{!}{%
\begin{tabular}{lcccccc}
\toprule
\textbf{Metric} & \textbf{Eval 1} & \textbf{Eval 2} & \textbf{Eval 3} & \textbf{Eval 4} & \textbf{LLM Eval} \\
\midrule
F1 score        & 0.750            & 0.774           & 0.702           & 0.748           & \textbf{0.778} \\
Pearson         & 0.970            & 0.975           & 0.945           & 0.990            & \textbf{0.997} \\
\bottomrule
\end{tabular}
}
\caption{Comparison of similarity metrics across results of the human (Eval 1, Eval 2, \ldots) and LLM  (LLM eval) evaluators against the results of the domain expert. The table shows the F1 scores (across individual insight comparisons) and Pearson correlation coefficients (of the final ELO score of all models) between evaluators and  domain expert on the private dataset.}
\label{tab:correlation}
\end{table}

To evaluate the capability of the LLM-based model, we also tasked four non-domain experts to solve the same task of comparing the same 100 insights of the private\_sales database. Then, the LLM evaluator (using the prompt from the Appendix~\ref{sec:llm_eval_prompt}) was tasked with the same objective, with the results in Figure~\ref{fig:insightfulness_llm_ci}. As seen in Table~\ref{tab:correlation}, the LLM evaluator reports better similarity with the domain expert both in the insight-to-insight F1 score and in the end-to-end Pearson correlation between Elo scores. Higher scores and correlations suggest better agreement with the  domain expert, thus the LLM evaluator can be used reliably to assess insightfulness.

Figure~\ref{fig:insightfulness_llm_ci_publicdbs} shows the results of the LLM evaluation of insightfulness on all public databases. Overall, all methods based on our approach perform better than baselines, with HLI clearly at the top and Quick performing the worst.

\subsection{Correctness Evaluation}
Following the correctness definition from Section~\ref{sec:insightcorrectness}, the truth value of each insight is the mean of the truth value of its claims scoring 1 when true and 0 when false. Some claims that contain multiple subclaims have been evaluated as $\frac{a}{b}$, where $b$ is the number of subclaims and $a$, the number of correct ones.\footnote{For example, the claim ``The top 5 products are Nafta, Natural, Diesel, Myt Vozidel, and Diesel'' has a truth value of $\frac{a}{5}$, where a is the number of correct products.} We consider the score of each method to be the mean correctness of the evaluated insights from that method.

\vspace{1ex}
\noindent{\bf Human Evaluation.} To evaluate correctness in the generated insights, a 
sample of 5 insights per database was hand-evaluated by the authors for each method (a total of 321 claims).


\begin{figure}[h]
    \includegraphics[width=0.5\textwidth]{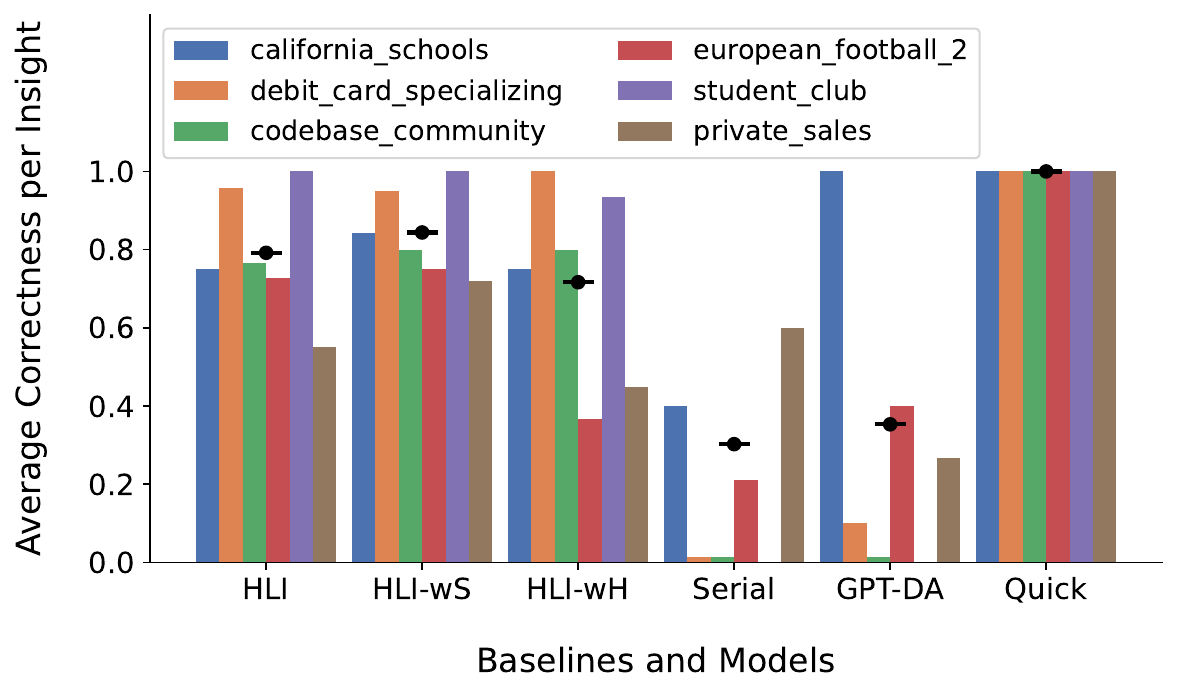}
    \caption{Average correctness per insight of all models across all databases. All evaluations with a total score of 0 correctness are due to having insights that have no truth value in relation to the data. For example an insight like the following: ``Identifying users who frequently switch roles can help in creating targeted engagement strategies.'' suggests potential action that cannot be evaluated with data from the database.}
    \label{fig:correctness_human}
\end{figure}

\begin{figure}[h]
    \includegraphics[width=0.5\textwidth]{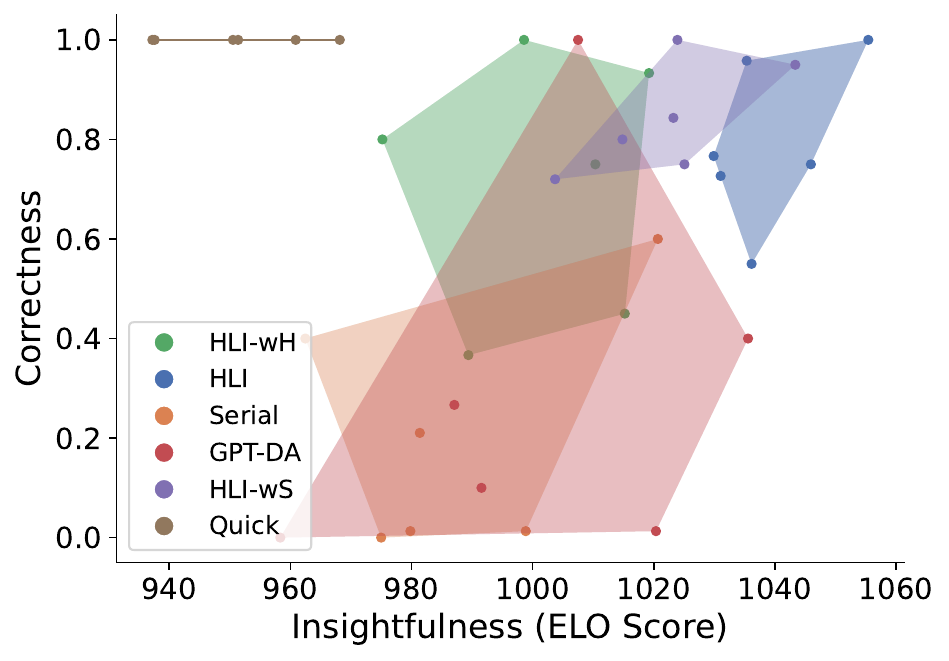}
    \caption{Dual evaluation of insightfulness (x-axis) and correctness (y-axis). Each polygon represents the performance variability of a model across different databases, with each point corresponding to the evaluation of one database. Larger polygon areas suggest greater variability across different domains and databases.}
    \label{fig:2axisplot}
\end{figure}

Figure~\ref{fig:correctness_human} illustrates the correctness evaluation across different databases for every method. Quick gets perfect correctness across all databases due to 
its design. {\sc HLI-wS} and {\sc HLI} show slightly lower results, but still consistently correct overall; {\sc HLI-wH} has a slightly lower correctness. Finally, Serial and GPT-DA get the lowest scores, with high variability. Overall, method {\sc HLI-wH} gives the best tradeoff between \textit{correctness} and \textit{insightfulness} over most databases. This can be seen in Figure~\ref{fig:2axisplot}, which represents the performance of each model on a database as a point in a plot, where the x-axis corresponds to the \textit{insightfulness} and the y-axis to \textit{correctness}. The methods are then represented as polygons, with a point representing the average performance, and an area representing the variability of 
its performance on different databases. Both {\sc HLI} and {\sc HLI-wS} have consistent results on the top-right part of the diagram, suggesting better performance, with {\sc HLI} being more insightful overall and {\sc HLI-wS} being slightly more correct. This aligns with the hypothesis that giving more information to the High-Level generator (as in {\sc HLI-wS}) constraints the space of possibilities, making questions that are easier to answer but less insightful.

Additionally, a manual review of 58 randomly sampled summaries and answers generated by our method revealed no hallucinations by the LLM when summarizing or transcribing the results of the generated SQL queries. The most common source of errors stemmed from semantic parsing issues.

\section{Related Work}
We first position our work w.r.t. the literature on insight generation. We then discuss Text-to-SQL and how 
it is relevant to our solution.

\vspace{1ex}\noindent
\textbf{Automatic Insights.}
Several models have been proposed for insight generation from tabular data, such as InsightPilot~\citep{ma2023demonstration}, QuickInsights~\citep{ding2019quickinsights}, OpenAI Data Analysis~\citep{openai2024data}, or Langchain's Pandas Agent~\citep{langchain2024}.

These works propose different definitions of insights. Some approaches consider insights derived from predefined templates~\citep{DBLP:journals/corr/abs-2008-13060} populated with aggregate measures across subsets of tables.
The generation of these 
insights is done either with a mining framework based on 
metrics~\citep{ding2019quickinsights} or using an LLM to guide the process~\citep{ma2023demonstration}. Although this approach offers several strengths, including providing visual data and ensuring factual correctness, it also presents certain limitations.
Other approaches~\citep{openai2024data, langchain2024} directly leverage LLMs to generate code to 
obtain insights. These methods are often specialized in single-step data analysis tasks (for example, calculating a correlation or doing some regression task) based on very concrete user instructions.
All 
previous approaches are based on one or more of the following assumptions and limitations:
\begin{itemize}
    \item Some approaches have only been tested on simple table structures and have not been evaluated on multi-table databases~\citep{ma2023demonstration, ding2019quickinsights, openai2024data, langchain2024}.
    \item Some approaches 
    require a user-defined goal (for example, ``Show me interesting trends in mathematics scores for students'') to guide the insight generation process~\citep{ma2023demonstration, ding2019quickinsights, openai2024data, langchain2024}. 
    \item Some approaches are template-based and require clean data that has been filtered and descriptive naming in columns~\citep{ding2019quickinsights}. For example, the insight ``[Subject]=[Math] has an increasing trend over [Time]'' becomes meaningless if the placeholders lack descriptive names or if they are irrelevant to the context, e.g., finding trends on IDs or telephone numbers.
\end{itemize}

\vspace{1ex}\noindent
\textbf{Text-to-SQL.}
Text-to-SQL is one of the most popular approaches for enabling natural language interfaces (NL) to databases. 
It  
consists of converting 
NL questions into valid SQL queries, 
where the result of the query is a set of tuples that answer the question.

The advent of 
LLMs 
has facilitated the development of 
complex architectures, such as LLM chains and agentic flows, which have significantly enhanced the efficacy of Text-to-SQL systems~\citep{tai-etal-2023-exploring, li2024petsql, gao2023texttosql, pourreza2023dinsql}. Our 
work employs such architectures to facilitate the automatic generation of insights from databases via automatically generated SQL queries.
While there is evidence that Text-to-SQL systems struggle with proprietary, large schemas~\citep{abs-2409-02038,papicchio2023qatch}, there are promising solutions to tackle this problem with prompt compression~\citep{li2024snapkv,tacl_a_00716}.

\section{Conclusions}

We have introduced a new method to generate insights from multi-table databases. We proposed an architecture that generates interesting questions, breaks them into more easily answerable subquestions, answers them with automatic SQL queries, and summarizes all results into a final insight. We evaluated the insightfulness and correctness of our method using human and LLM evaluation against three baselines and two variations. We demonstrated that our method achieves better and more consistent results on these two metrics.

For future research, the model's evaluation could be extended to encompass a broader range of domains or data sources (e.g., textual documentation using RAG, Knowledge Graphs or Internet search). Also, there are techniques that might enhance the performance, such as k-shot learning or prompt learning techniques~\citep{yuksekgonul2024textgradautomaticdifferentiationtext}. Finally, an interactive exploration of the insight search space could lead to better outputs with a limited effort, for example by extending the solution to consider user feedback on early results.

\section{Limitations}
Generated insights are not always perfectly correct, which in some contexts may lead to unexpected decisions. This effect can be alleviated by reducing database complexity or with future models that have better results on the Text-to-SQL task.
Additionally, the architecture of the model involves a lot of LLM calls, which can have downsides in terms of cost (see Appendix~\ref{sec:cost} for more details).
Although different domains have been evaluated, our effort is still limited by cost and some domains have not been covered. Finally, only English insights have been generated and our architecture has only been evaluated with one LLM.

\section{Ethical Considerations}
The use of LLMs for generating automatic data insights presents several ethical challenges.

A primary concern in all works that use LLMs is the potential risk of biases, as large datasets may inadvertently reinforce societal prejudices. We also need to account for its environmental impact, due to the high computational demands of LLMs.

There is also a risk of misuse of our system, as generated insights with our framework may be exploited for unethical purposes, such as manipulation or surveillance.

Another risk of our system is its implementation in real-life scenarios. Automated systems should not be used to deflect responsibility in decision-making contexts, they should be used as a supporting tool. Finally, there is a risk of job displacement, as reliance on AI may reduce the need for human analysts in the long term.

\section{Acknowledgements}
We thank the action editor and the reviewers for their comments that helped us improve the content of our work. We also thank all the people from Aily who have helped us, especially Roumila Chebil and the genAI team (Gerard Conangla, Nekane Guarrotxena, Marco Introvigne, Miguel Conner, Fernando Rodríguez, Juan Abia and Joshua Fellowes).

\appendix

\bibliographystyle{acl_natbib.bst}
\bibliography{references}

\begin{thebibliography}{39}
\expandafter\ifx\csname natexlab\endcsname\relax\def\natexlab#1{#1}\fi

\bibitem[{Abdul-Azeez et~al.(2024)Abdul-Azeez, Ihechere, and Idemudia}]{oluwatosin_abdulazeez__2024}
Oluwatosin Abdul-Azeez, Alexsandra~Ogadimma Ihechere, and Courage Idemudia. 2024.
\newblock \href {https://doi.org/10.51594/ijmer.v6i7.1257} {Enhancing business performance: The role of data-driven analytics in strategic decision-making}.
\newblock \emph{International journal of management and entrepreneurship research}, 6(7):2066--2081.

\bibitem[{Arora and Malik(2015)}]{deepali_arora__2015}
Deepali Arora and Piyush Malik. 2015.
\newblock \href {https://doi.org/10.1109/BIGDATASERVICE.2015.62} {Analytics: Key to go from generating big data to deriving business value}.

\bibitem[{Askell et~al.(2021)Askell, Bai, Chen, Drain, Ganguli, Henighan, Jones, Joseph, Mann, DasSarma, Elhage, Hatfield-Dodds, Hernandez, Kernion, Ndousse, Olsson, Amodei, Brown, Clark, McCandlish, Olah, and Kaplan}]{askell2021general}
Amanda Askell, Yuntao Bai, Anna Chen, Dawn Drain, Deep Ganguli, Tom Henighan, Andy Jones, Nicholas Joseph, Ben Mann, Nova DasSarma, Nelson Elhage, Zac Hatfield-Dodds, Danny Hernandez, Jackson Kernion, Kamal Ndousse, Catherine Olsson, Dario Amodei, Tom Brown, Jack Clark, Sam McCandlish, Chris Olah, and Jared Kaplan. 2021.
\newblock \href {http://arxiv.org/abs/2112.00861} {A general language assistant as a laboratory for alignment}.

\bibitem[{Bean(2022)}]{bean2022data}
Randy Bean. 2022.
\newblock Why becoming a data-driven organization is so hard.
\newblock \emph{Harvard Business Review}.
\newblock Available online: https://hbr.org/2022/02/why-becoming-a-data-driven-organization-is-so-hard.

\bibitem[{Boubdir et~al.(2023)Boubdir, Kim, Ermis, Hooker, and Fadaee}]{boubdir2023elo}
Meriem Boubdir, Edward Kim, Beyza Ermis, Sara Hooker, and Marzieh Fadaee. 2023.
\newblock \href {http://arxiv.org/abs/2311.17295} {Elo uncovered: Robustness and best practices in language model evaluation}.

\bibitem[{Chen et~al.(2024)Chen, Wenz, Zhang, Kayali, Tatbul, Cafarella, Demiralp, and Stonebraker}]{abs-2409-02038}
Peter~Baile Chen, Fabian Wenz, Yi~Zhang, Moe Kayali, Nesime Tatbul, Michael~J. Cafarella, {\c{C}}agatay Demiralp, and Michael Stonebraker. 2024.
\newblock \href {https://doi.org/10.48550/ARXIV.2409.02038} {{BEAVER:} an enterprise benchmark for text-to-sql}.
\newblock \emph{CoRR}, abs/2409.02038.

\bibitem[{Chiang et~al.(2024)Chiang, Zheng, Sheng, Angelopoulos, Li, Li, Zhang, Zhu, Jordan, Gonzalez, and Stoica}]{chiang2024chatbot}
Wei-Lin Chiang, Lianmin Zheng, Ying Sheng, Anastasios~Nikolas Angelopoulos, Tianle Li, Dacheng Li, Hao Zhang, Banghua Zhu, Michael Jordan, Joseph~E. Gonzalez, and Ion Stoica. 2024.
\newblock \href {http://arxiv.org/abs/2403.04132} {Chatbot arena: An open platform for evaluating llms by human preference}.

\bibitem[{Corallo and Papotti(2024)}]{tacl_a_00716}
Giulio Corallo and Paolo Papotti. 2024.
\newblock \href {https://doi.org/10.1162/tacl_a_00716} {Finch: Prompt-guided key-value cache compression for large language models}.
\newblock \emph{Transactions of the Association for Computational Linguistics}, 12:1517--1532.

\bibitem[{{Creative Commons}(2013)}]{cc_by_sa_4_0}
{Creative Commons}. 2013.
\newblock Attribution-sharealike 4.0 international (cc by-sa 4.0).
\newblock \url{https://creativecommons.org/licenses/by-sa/4.0/}.

\bibitem[{Daraojimba et~al.(2024)Daraojimba, Victoria, Ibeh, Elufioye, Olorunsogo, Asuzu, and Nduubuisi}]{andrew_ifesinachi_daraojimba__2024}
Andrew~Ifesinachi Daraojimba, Chidera Victoria, Ibeh, Oluwafunmi~Adijat Elufioye, Temidayo Olorunsogo, Onyeka~Franca Asuzu, and Ndubuisi~Leonard Nduubuisi. 2024.
\newblock \href {https://doi.org/10.30574/wjarr.2024.21.2.0246} {Data analytics in healthcare: A review of patient-centric approaches and healthcare delivery}.
\newblock \emph{World Journal Of Advanced Research and Reviews}.

\bibitem[{Ding et~al.(2019)Ding, Han, Xu, Zhang, and Zhang}]{ding2019quickinsights}
Rui Ding, Shi Han, Yong Xu, Haidong Zhang, and Dongmei Zhang. 2019.
\newblock \href {https://www.microsoft.com/en-us/research/publication/quickinsights-quick-and-automatic-discovery-of-insights-from-multi-dimensional-data/} {Quickinsights: Quick and automatic discovery of insights from multi-dimensional data}.
\newblock In \emph{Proceedings of the 2019 ACM International Conference on Management of Data (SIGMOD'19 industrial track)}.

\bibitem[{Gao et~al.(2023)Gao, Wang, Li, Sun, Qian, Ding, and Zhou}]{gao2023texttosql}
Dawei Gao, Haibin Wang, Yaliang Li, Xiuyu Sun, Yichen Qian, Bolin Ding, and Jingren Zhou. 2023.
\newblock \href {http://arxiv.org/abs/2308.15363} {Text-to-sql empowered by large language models: A benchmark evaluation}.

\bibitem[{Jahns(2013)}]{veit_jahns_2013}
Veit Jahns. 2013.
\newblock \href {https://doi.org/10.1145/2532780.2544424} {Data insights: new ways to visualize and make sense of data by hunter whitney}.
\newblock \emph{ACM Sigsoft Software Engineering Notes}, 38(6):45--46.

\bibitem[{Khan(2024)}]{attia_khan_2024}
Attia Khan. 2024.
\newblock \href {https://doi.org/10.55041/ijsrem32598} {Effective decision making using data analytics}.
\newblock \emph{Indian Scientific Journal Of Research In Engineering And Management}, 08(04):1--5.

\bibitem[{LangChain(2024{\natexlab{a}})}]{langchain_sql_agent}
LangChain. 2024{\natexlab{a}}.
\newblock Langgraph sql agent.
\newblock \url{https://langchain-ai.github.io/langgraph/tutorials/sql-agent/}.

\bibitem[{LangChain(2024{\natexlab{b}})}]{langchain2024}
LangChain. 2024{\natexlab{b}}.
\newblock Pandas dataframe.
\newblock \url{https://python.langchain.com/v0.2/docs/integrations/toolkits/pandas/}.

\bibitem[{Law et~al.(2020)Law, Endert, and Stasko}]{DBLP:journals/corr/abs-2008-13060}
Po{-}Ming Law, Alex Endert, and John~T. Stasko. 2020.
\newblock \href {http://arxiv.org/abs/2008.13060} {Characterizing automated data insights}.
\newblock \emph{CoRR}, abs/2008.13060.

\bibitem[{Li et~al.(2023)Li, Hui, Qu, Yang, Li, Li, Wang, Qin, Cao, Geng, Huo, Zhou, Ma, Li, Chang, Huang, Cheng, and Li}]{li2023llm}
Jinyang Li, Binyuan Hui, Ge~Qu, Jiaxi Yang, Binhua Li, Bowen Li, Bailin Wang, Bowen Qin, Rongyu Cao, Ruiying Geng, Nan Huo, Xuanhe Zhou, Chenhao Ma, Guoliang Li, Kevin C.~C. Chang, Fei Huang, Reynold Cheng, and Yongbin Li. 2023.
\newblock \href {http://arxiv.org/abs/2305.03111} {Can llm already serve as a database interface? a big bench for large-scale database grounded text-to-sqls}.

\bibitem[{Li et~al.(2024{\natexlab{a}})Li, He, Yashar, Cui, Ge, Zhang, Rifinski~Fainman, Zhang, and Chaudhuri}]{10.1145/3654979}
Peng Li, Yeye He, Dror Yashar, Weiwei Cui, Song Ge, Haidong Zhang, Danielle Rifinski~Fainman, Dongmei Zhang, and Surajit Chaudhuri. 2024{\natexlab{a}}.
\newblock \href {https://doi.org/10.1145/3654979} {Table-gpt: Table fine-tuned gpt for diverse table tasks}.
\newblock \emph{Proc. ACM Manag. Data}, 2(3).

\bibitem[{Li et~al.(2024{\natexlab{b}})Li, Huang, Yang, Venkitesh, Locatelli, Ye, Cai, Lewis, and Chen}]{li2024snapkv}
Yuhong Li, Yingbing Huang, Bowen Yang, Bharat Venkitesh, Acyr Locatelli, Hanchen Ye, Tianle Cai, Patrick Lewis, and Deming Chen. 2024{\natexlab{b}}.
\newblock \href {https://openreview.net/forum?id=poE54GOq2l} {Snap{KV}: {LLM} knows what you are looking for before generation}.
\newblock In \emph{The Thirty-eighth Annual Conference on Neural Information Processing Systems}.

\bibitem[{Li et~al.(2024{\natexlab{c}})Li, Wang, Zhao, Yang, Du, Hu, Zhang, Ye, Li, Zhao, and Mao}]{li2024petsql}
Zhishuai Li, Xiang Wang, Jingjing Zhao, Sun Yang, Guoqing Du, Xiaoru Hu, Bin Zhang, Yuxiao Ye, Ziyue Li, Rui Zhao, and Hangyu Mao. 2024{\natexlab{c}}.
\newblock \href {http://arxiv.org/abs/2403.09732} {Pet-sql: A prompt-enhanced two-stage text-to-sql framework with cross-consistency}.

\bibitem[{Lin and Chen(2023)}]{lin2023llmevalunifiedmultidimensionalautomatic}
Yen-Ting Lin and Yun-Nung Chen. 2023.
\newblock \href {http://arxiv.org/abs/2305.13711} {Llm-eval: Unified multi-dimensional automatic evaluation for open-domain conversations with large language models}.

\bibitem[{Liu et~al.(2023)Liu, Iter, Xu, Wang, Xu, and Zhu}]{liu2023gevalnlgevaluationusing}
Yang Liu, Dan Iter, Yichong Xu, Shuohang Wang, Ruochen Xu, and Chenguang Zhu. 2023.
\newblock \href {http://arxiv.org/abs/2303.16634} {G-eval: Nlg evaluation using gpt-4 with better human alignment}.

\bibitem[{LMSys(May 22, 2023)}]{GoogleColab2024}
LMSys. May 22, 2023.
\newblock Chatbot arena: Elo rating calculation.
\newblock \url{https://colab.research.google.com/drive/17L9uCiAivzWfzOxo2Tb9RMauT7vS6nVU?usp=sharing}.

\bibitem[{Ma et~al.(2023)Ma, Ding, Wang, Han, and Zhang}]{ma2023demonstration}
Pingchuan Ma, Rui Ding, Shuai Wang, Shi Han, and Dongmei Zhang. 2023.
\newblock \href {http://arxiv.org/abs/2304.00477} {Demonstration of insightpilot: An llm-empowered automated data exploration system}.

\bibitem[{{Microsoft}()}]{powerbi}
{Microsoft}.
\newblock Power bi.
\newblock \url{https://powerbi.microsoft.com}.

\bibitem[{OpenAI(2024{\natexlab{a}})}]{openai2024data}
OpenAI. 2024{\natexlab{a}}.
\newblock \href {https://help.openai.com/en/articles/8437071-data-analysis-with-chatgpt} {Data analysis with chatgpt}.
\newblock \url{https://help.openai.com/en/articles/8437071-data-analysis-with-chatgpt}.

\bibitem[{OpenAI(2024{\natexlab{b}})}]{openai_gpt4o_2024}
OpenAI. 2024{\natexlab{b}}.
\newblock Hello gpt-4o.
\newblock \url{https://openai.com/index/hello-gpt-4o/}.

\bibitem[{{OpenAI}(2024)}]{openai2024}
{OpenAI}. 2024.
\newblock Openai function calling.
\newblock \url{https://platform.openai.com/docs/guides/function-calling}.

\bibitem[{Papicchio et~al.(2023)Papicchio, Papotti, and Cagliero}]{papicchio2023qatch}
Simone Papicchio, Paolo Papotti, and Luca Cagliero. 2023.
\newblock \href {https://openreview.net/forum?id=XOpaPrb0U5} {{QATCH}: Benchmarking {SQL}-centric tasks with table representation learning models on your data}.
\newblock In \emph{Thirty-seventh Conference on Neural Information Processing Systems Datasets and Benchmarks Track}.

\bibitem[{Pawar et~al.(2024)Pawar, Tonmoy, Zaman, Jain, Chadha, and Das}]{pawar2024whatwhycontextlength}
Saurav Pawar, S.~M Towhidul~Islam Tonmoy, S~M~Mehedi Zaman, Vinija Jain, Aman Chadha, and Amitava Das. 2024.
\newblock \href {http://arxiv.org/abs/2401.07872} {The what, why, and how of context length extension techniques in large language models -- a detailed survey}.

\bibitem[{Pourreza and Rafiei(2023)}]{pourreza2023dinsql}
Mohammadreza Pourreza and Davood Rafiei. 2023.
\newblock \href {http://arxiv.org/abs/2304.11015} {Din-sql: Decomposed in-context learning of text-to-sql with self-correction}.

\bibitem[{Rajkumar et~al.(2022)Rajkumar, Li, and Bahdanau}]{rajkumar2022evaluating}
Nitarshan Rajkumar, Raymond Li, and Dzmitry Bahdanau. 2022.
\newblock \href {http://arxiv.org/abs/2204.00498} {Evaluating the text-to-sql capabilities of large language models}.

\bibitem[{Shinn et~al.(2023)Shinn, Cassano, Berman, Gopinath, Narasimhan, and Yao}]{shinn2023reflexionlanguageagentsverbal}
Noah Shinn, Federico Cassano, Edward Berman, Ashwin Gopinath, Karthik Narasimhan, and Shunyu Yao. 2023.
\newblock \href {http://arxiv.org/abs/2303.11366} {Reflexion: Language agents with verbal reinforcement learning}.

\bibitem[{Steiner(2022)}]{stefan_steiner_2022}
Stefan Steiner. 2022.
\newblock \href {https://doi.org/10.33548/scientia768} {Harnessing data to make better-informed decisions}.
\newblock \emph{Scientia}.

\bibitem[{Sun et~al.(2024)Sun, Arik, Muzio, Miculicich, Gundabathula, Yin, Dai, Nakhost, Sinha, Wang, and Pfister}]{sun2024sqlpalmimprovedlargelanguage}
Ruoxi Sun, Sercan~Ö. Arik, Alex Muzio, Lesly Miculicich, Satya Gundabathula, Pengcheng Yin, Hanjun Dai, Hootan Nakhost, Rajarishi Sinha, Zifeng Wang, and Tomas Pfister. 2024.
\newblock \href {http://arxiv.org/abs/2306.00739} {Sql-palm: Improved large language model adaptation for text-to-sql (extended)}.

\bibitem[{Tai et~al.(2023)Tai, Chen, Zhang, Deng, and Sun}]{tai-etal-2023-exploring}
Chang-Yu Tai, Ziru Chen, Tianshu Zhang, Xiang Deng, and Huan Sun. 2023.
\newblock \href {https://doi.org/10.18653/v1/2023.emnlp-main.327} {Exploring chain of thought style prompting for text-to-{SQL}}.
\newblock In \emph{Proceedings of the 2023 Conference on Empirical Methods in Natural Language Processing}, pages 5376--5393, Singapore. Association for Computational Linguistics.

\bibitem[{Volvovsky et~al.(2024)Volvovsky, Marcassa, and Panbiharwala}]{volvovsky2024dfinsqlintegratingfocusedschema}
Shai Volvovsky, Marco Marcassa, and Mustafa Panbiharwala. 2024.
\newblock \href {http://arxiv.org/abs/2403.00872} {Dfin-sql: Integrating focused schema with din-sql for superior accuracy in large-scale databases}.

\bibitem[{Yuksekgonul et~al.(2024)Yuksekgonul, Bianchi, Boen, Liu, Huang, Guestrin, and Zou}]{yuksekgonul2024textgradautomaticdifferentiationtext}
Mert Yuksekgonul, Federico Bianchi, Joseph Boen, Sheng Liu, Zhi Huang, Carlos Guestrin, and James Zou. 2024.
\newblock \href {http://arxiv.org/abs/2406.07496} {Textgrad: Automatic "differentiation" via text}.

\end{thebibliography}

\section{Example of an Insight}
Generated by High-Level Insights

\textbf{Insight}:
Higher percentages of students eligible for free or reduced-price meals (FRPM) correlate with lower average SAT scores in reading, math, and writing. Charter schools, which have a higher average FRPM eligibility (61.07\%), also show lower SAT scores compared to non-charter schools. This indicates a strong link between socioeconomic status and academic performance.

\textbf{High-Level Question}:
What are the trends in academic performance across different socioeconomic groups?

\textbf{Subquestions}:
\begin{itemize}
    \item Are there significant differences in SAT performance between charter schools and non-charter schools, particularly in relation to their students' socioeconomic status?
    \item How do the average SAT scores in reading, math, and writing vary across different counties?
    \item For schools with a high percentage of students eligible for free or reduced-price meals (>70\%), what is the average number of students scoring 1500 or above on the SAT?
    \item What is the correlation between the percentage of students eligible for free or reduced-price meals and the average SAT scores in reading, math, and writing?
\end{itemize}

\section{Dataset Details}

\subsection{Dataset License}
The generated dataset of insights is licensed under CC BY-SA 4.0~\citep{cc_by_sa_4_0}.
This dataset is intended for research purposes.

The released dataset does not contain any personal or confidential information and is fully derived from public data~\citep{li2023llm}. All generated insights are in English.

\subsection{Number of insights}
\label{sec:insight_number}

\begin{table}[H]
\centering
\renewcommand{\arraystretch}{2.3}
\resizebox{0.48\textwidth}{!}{%
\begin{tabular}{lcccccc}
\toprule
\textbf{Model} & \textbf{HLI} & \textbf{HLI-wS} & \textbf{HLI-wH} & \textbf{Serial} & \textbf{GPT-DA} & \textbf{Quick} \\
\midrule
\makecell{california\_ \\ schools} & 7 & 9 & 10 & 9 & 9 & 10 \\
\makecell{codebase\_ \\ community} & 9 & 6 & 9 & 9 & 10 & 6 \\
\makecell{debit\_card\_ \\ specializing} & 7 & 5 & 10 & 9 & 10 & 10 \\
\makecell{european\_ \\ football\_2} & 7 & 5 & 6 & 9 & 10 & 10 \\
\makecell{private\_ \\ sales} & 6 & 6 & 7 & 9 & 10 & 9 \\
\makecell{student\_ \\ club} & 6 & 6 & 10 & 9 & 10 & 10 \\
\bottomrule
\end{tabular}
}
\caption{Number of insights generated per database for each method.}
\label{tab:insight_num}
\end{table}

Table~\ref{tab:insight_num} shows that the number of generated insights varies from method to method because of different reasons:
\begin{itemize}
    \item {\sc HLI}, {\sc HLI-wS} and {\sc HLI-wH}: Because of the LLM evaluation functions which filter final insights if they get a low score.
    \item GPT-DA: In one instance the model only generated 9 insights even when explicitly it was told to generate 10.
    \item Quick: The model was either impossible to pre-join (private\_sales) or the pre-joined table only generated 1 number (codebase\_community). In both cases, insights were drawn from randomly sampled individual tables.
\end{itemize}

\subsection{Number of subquestions}

\begin{table}[H]
\centering
\renewcommand{\arraystretch}{2.3}
\resizebox{0.48\textwidth}{!}{%
\begin{tabular}{lcccccc}
\toprule
\textbf{Model} & \textbf{HLI} & \textbf{HLI-wS} & \textbf{HLI-wH} & \textbf{Serial} & \textbf{GPT-DA} & \textbf{Quick} \\
\midrule
\makecell{california\_ \\ schools} & 3.71 & 4.78 & 1.00 & 1.00 & 1.00 & 1.00 \\
\makecell{codebase\_ \\ community} & 3.89 & 1.83 & 1.00 & 1.00 & 1.00 & 1.00 \\
\makecell{debit\_card\_ \\ specializing} & 4.00 & 2.20 & 1.00 & 1.00 & 1.00 & 1.00 \\
\makecell{european\_ \\ football\_2} & 2.14 & 3.00 & 1.00 & 1.00 & 1.00 & 1.00 \\
\makecell{private\_ \\ sales} & 2.00 & 1.83 & 1.00 & 1.00 & 1.00 & 1.00 \\
\makecell{student\_ \\ club} & 3.00 & 3.50 & 1.00 & 1.00 & 1.00 & 1.00 \\
\bottomrule
\end{tabular}
}
\caption{Average number of subquestions generated for each insight}
\label{tab:insight_subq}
\end{table}

The average number of subquestions generated by each insight is presented in Table~\ref{tab:insight_subq}.

\subsection{Insight length}

\begin{table}[H]
\centering
\renewcommand{\arraystretch}{2.3}
\resizebox{0.48\textwidth}{!}{%
\begin{tabular}{lcccccc}
\toprule
\textbf{Model} & \textbf{HLI} & \textbf{HLI-wS} & \textbf{HLI-wH} & \textbf{Serial} & \textbf{GPT-DA} & \textbf{Quick} \\
\midrule
\makecell{california\_ \\ schools} & 424.0 & 412.0 & 270.2 & 207.89 & 226.78 & 89.0 \\
\makecell{codebase\_ \\ community} & 349.11 & 298.5 & 177.78 & 271.44 & 199.0 & 87.0 \\
\makecell{debit\_card\_ \\ specializing} & 438.86 & 332.8 & 213.6 & 246.89 & 218.7 & 104.6 \\
\makecell{european\_ \\ football\_2} & 297.57 & 323.4 & 217.33 & 307.78 & 211.1 & 60.9 \\
\makecell{private\_ \\ sales} & 423.83 & 314.0 & 242.29 & 350.33 & 99.0 & 90.56 \\
\makecell{student\_ \\ club} & 270.0 & 257.83 & 118.10 & 296.22 & 471.67 & 55.25 \\
\bottomrule
\end{tabular}
}
\caption{Average number of characters per insight}
\label{tab:insight_len}
\end{table}

The average length in characters for each insight is presented in Table~\ref{tab:insight_len}.

\section{Evaluation Details}
\subsection{Instructions}

\begin{tcolorbox}[colback=gray!23!white, colframe=black, title=Instructions for the evaluators, breakable]

You will receive a list containing pairs of insights. Your task is to decide which insight in each pair is more insightful based on the provided insightfulness criteria:\\

Insightfulness measures the quality of an insight. More concretely, it measures criteria like
\begin{itemize}
    \item How impactful/actionable the insight is, if it can lead to any key actions and what the impact of those actions would be (ideally over measurable KPIs)
    \item How interesting and relevant is the information provided.
    \item How clear, precise and coherent the insight is, how easily and accurately can you understand the information it provides
    \item \ldots
\end{itemize}

All these criteria are subjective and may vary based on your expertise, domain, and specific interests. You'll need to use your judgement to assess which insight offers more value based on these criteria.\\

Evaluation Process
\begin{enumerate}
    \item Read both insights
    \item Compare the insights and choose the more insightful one
    \item Write your chosen insight in the Best\_Insight column. Write ``1'' if you choose the insight to the left and ``2'' if you choose the one to the right
\end{enumerate}

Additional Guidelines
\begin{itemize}
    \item Correctness: The correctness of insights will be evaluated separately. Do not factor correctness into your evaluation.
    \item Length Variability: Insights may vary in length.
\end{itemize}
\end{tcolorbox}

\subsection{Evaluators Profile}
\subsubsection{Recruitment}
All evaluators were recruited from our company.

\subsubsection{Data Statement}
The following is a self-reported data statement.

\begin{table}[h]
\centering
\renewcommand{\arraystretch}{2}
\resizebox{0.48\textwidth}{!}{%
\begin{tabular}{lcccc}
\toprule
\textbf{Evaluator} & \textbf{Age} & \textbf{Gender} & \makecell{\textbf{Nationality}} & \makecell{\textbf{Professional} \\ \textbf{Background}} \\
\midrule
\makecell{Expert \\ Eval} & 40-50 & Female & \makecell{Algeria/ \\ Germany} & \makecell{Business \\ Impact Lead} \\
\makecell{Eval 1} & 30-40 & Male & Spain & \makecell{AI \\ Engineer} \\
\makecell{Eval 2} & 20-30 & Female & Spain & \makecell{AI \\ Engineer} \\
\makecell{Eval 3} & 20-30 & Male & Italy & \makecell{AI \\ Engineer} \\
\makecell{Eval 4} & 20-30 & Male & USA & \makecell{AI \\ Engineer} \\
\bottomrule
\end{tabular}
}
\caption{Summary of Evaluators}
\end{table}

\section{ELO K-Factor Study}
\label{section:kfactor}

\begin{figure*}[h]
\includegraphics[width=\textwidth]{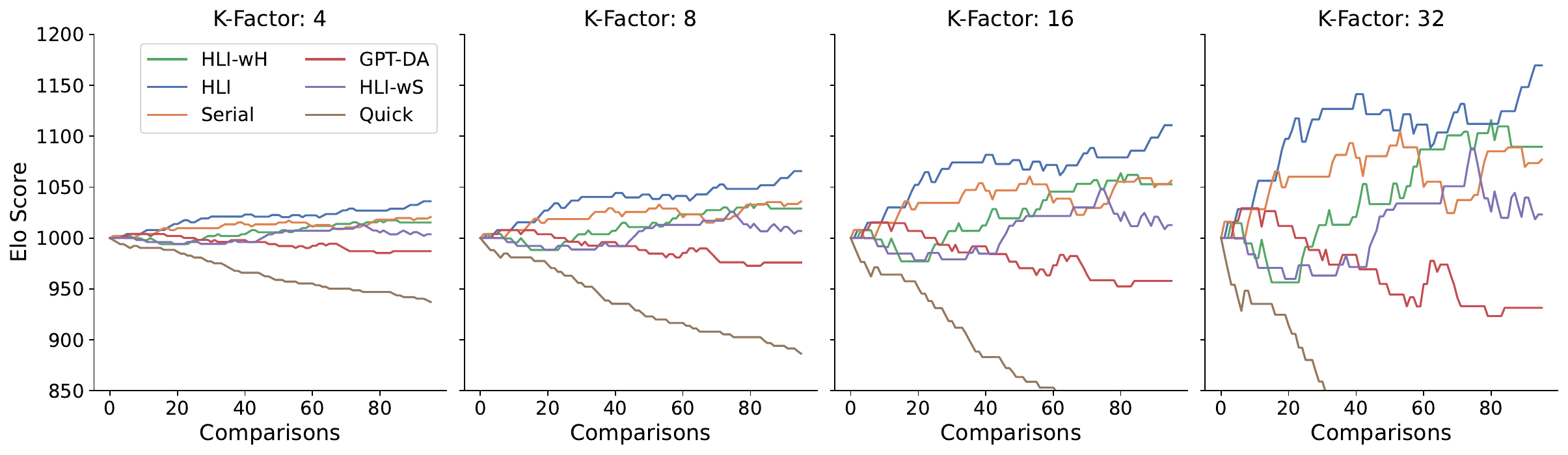}
    \caption{Study on the effect of the k-factor on the ELO evolution. We can see that 100 comparisons are enough to grant convergence and stability. However, greater k-factors like 16 or 32 give more weight to the latest comparisons (giving a better last ELO score to HLI-wH over Serial just for winning the latest comparisons).}
    \label{fig:k-factor-studdy}
\end{figure*}

Figure~\ref{fig:k-factor-studdy} shows a study on the effect of the k-factor hyperparameter on the ELO evaluation of our domain expert. We can see that higher k-factors provide faster but less stable convergence, with more weight of the latest comparisons. With a k-factor of 4 or 8, our study suggests 100 comparisons are enough to converge into a final result. We can see that a k-factor of 16 or 32 gives different final ELO results (HLI-wH gets a better position than GPT-DA) due to the increased weight of latest comparisons.

\section{Prompts}

\subsection{Database Description}
\label{sec:dbdesc_prompt}

\begin{tcolorbox}[colback=gray!5!white, colframe=black, title=Prompt 1: Database Description Creation Prompt, breakable]

Given the following database create a description of the database in natural text explaining to a user the structure of the table and the data it contains. Avoid html and give raw text with all the explanations. Explain the database, the tables and the columns in each table.

\{db\_schema\}

\{db\_sample\_rows\}

\end{tcolorbox}

\subsection{High-Level Insights Prompts}
\label{sec:hli_prompt}

\begin{tcolorbox}[colback=gray!5!white, colframe=black, title=Prompt 2: High-Level Generator Prompt, breakable]

Generate ten questions related to a user that works as an analyst from the following data sources: given a retail worker who has to analyze customer data - what are the customer trends, and what actions might they need to take using data sources. The user works with the following database: \{tables\_description\}
\end{tcolorbox}

\begin{tcolorbox}[colback=gray!5!white, colframe=black, title=Prompt 3: Low-level Generator Prompt, breakable]

We have a platform for users that need to analyse data.\\
We have access to these sql tables \{tables\}. The tables have the following description \{tables\_description\}. We have the following complex questions \{questions\}.\\
Generate subquestions that a user can use to answer the complex question which might be answered from these tables. The questions should
be answered in the form of insights that can be used to make decisions not just information about some numbers.The questions should be very verbose that if sum is needed say sum and if average is needed say average of certain columns. The
questions should non sequential and can be  executed in parallel. You can use the
following schema as a reference: \{schema\}

\end{tcolorbox}




\subsection{High-Level Insight without High-Level ({\sc HLI-wH})}
\label{sec:hli-wh_prompt}

\begin{tcolorbox}[colback=gray!5!white, colframe=black, title=Prompt 4: Hypothesis Generator prompt, breakable]

We have a platform for users that need to analyse data.\\
We have access to these sql tables \{tables\}.
The tables have the following description \{tables\_description\}.\\
Generate ten questions related to customer consumption data in the industry for the user from the following data sources: The questions should be answered in the form of insights that can be used to make decisions not just information about some numbers.\\
The questions should be very verbose that if sum is needed say sum and if average is needed say average of certain columns. The questions should be non sequential and can be executed in parallel. You can use the following schema as a reference: \{schema\}

\end{tcolorbox}

\subsection{Baselines}
\label{sec:baseline_prompt}

\begin{tcolorbox}[colback=gray!5!white, colframe=black, title=Prompt 4: Hypothesis Generator prompt, breakable]

We have a platform for users that need to analyse data.\\
We have access to these sql tables \{tables\}.
The tables have the following description \{tables\_description\}.\\
Generate ten insights (3 lines long) related to customer consumption data in the industry for the user from the following data sources: The insights must be used to make decisions not just information about some numbers.

You can use the following schema as a reference: \{schema\}

Data: \{Data\}

\end{tcolorbox}

\subsection{Short Database Description}
\label{sec:short_dbdesc_prompt}

\begin{tcolorbox}[colback=gray!5!white, colframe=black, title=Prompt 5: Hypothesis Generator prompt, breakable]

Take the following database description and return a brief (3 lines) description.

Your description must contain only a high-level description of the database, avoid explaining tables.

\#\#\#

Database description: \{tables\_description\}

\end{tcolorbox}

\subsection{SQL Agent}
\label{sec:sqlagent_prompt}

The following prompt is from the langgraph sql agent: ~\citep{langchain_sql_agent}

\begin{tcolorbox}[colback=gray!5!white, colframe=black, title=Prompt 6: SQL Agent Prompt from Langgraph, breakable]

You are an agent designed to interact with a SQL database.

Given an input question, create a syntactically correct \{dialect\} query to run, then look at the results of the query and return the answer

Unless the user specifies a specific number of examples they wish to obtain, always limit your query to at most \{top\_k\} results.

You can order the results by a relevant column to return the most interesting examples in the database.

Never query for all the columns from a specific table, only ask for the relevant columns given the question.

You have access to tools for interacting with the database.

Only use the below tools. Only use the information returned by the below tools to construct your final answer.

You MUST double check your query before executing it. If you get an error while executing a query, rewrite the query and try again.

DO NOT make any DML statements (INSERT, UPDATE, DELETE, DROP etc.) to the database.

If the question does not seem related to the database, just return ``I don't know'' as the answer.

\end{tcolorbox}

\subsection{Summarization Prompts}
\label{sec:summarization_prompt}

\begin{tcolorbox}[colback=gray!5!white, colframe=black, title=Prompt 7: Summarizer prompt, breakable]

As an AI programmed to simulate an expert-level business analyst, your task is to construct a short strategic business insight from provided data.\\
  
Could you please provide a concise and comprehensive summary of the given text? The summary should capture the main
points and key details of the text while conveying the author's intended meaning accurately. The length of the summary
should be around 3 lines, gathering the main points and representing it in 3 lines.\\

The final insight should seem realistic and actionable, serving as a powerful tool for decision-makers to visualize potential strategies and outcomes.\\
  
Your summary must follow this guidelines
\begin{itemize}
    \item Don't write insights longer than 3 lines long.
    \item Avoid adding information or recommendations that don't come from the context.
    \item Avoid enumeration of facts; reason the high-level pattern from the data and most important information.
    \item Mention the important data to support the inferred high-level patterns.
    \item Avoid focusing on individual entities, unless it's for exemplifying a pattern.
    \item Provide information that is actionable, not just random facts; it must only provide the interesting information.
    \item Avoid mentioning unimportant information.
    \item You don't need to mention all the information from the data, only the important one.
\end{itemize}
Remember to follow this instructions (after doing the summary, recheck that your summary followings them) or I will lose my job :(\\
  
High-Level Question\\
\{hlquestion\}\\

Context\\
\{low\_level\_answers\}\\

\end{tcolorbox}

\begin{tcolorbox}[colback=gray!5!white, colframe=black, title=Prompt 8: Summarization with reflection prompt, breakable]

Please reflect on your recent summarization task. You will be provided with the context and corresponding summary, alongside with the score that the summary received and its reasoning. You will need to produce a new summarized strategic business insight from provided data, considering the score and the reasoning to improve the result.
Remember that the final insight should seem realistic and actionable, serving as a powerful tool for decision-makers to visualize
potential strategies and outcomes.\\

Avoid adding extra information or recommendations.
Focus on the data coming from the context, but don't just enumerate it, reason and provide an actionable insight for the user with the given task using the information from the context.\\

\#\#\#\\

Context:\\
\{low\_level\_answers\}\\

\#\#\#\\

Summary:\\
\{summary\}\\

\#\#\#\\

Score:\\
This summary was evaluated with a score of: \{score\}\\

Reasoning:\\
The reasons for the score:
\{reasoning\}\\

\#\#\#\\

Guidelines for a Good Summary:
\begin{itemize}
    \item Identify Key Points: Focus on the main ideas and essential details of the original text.
    \item Be Concise: Use clear and concise language to convey the information.
    \item Avoid Personal Opinions: Ensure the summary is objective and free from personal bias.
    \item Use Your Own Words: Paraphrase the original text to avoid plagiarism.
    \item Maintain Coherence: Ensure the summary is well-organized and flows logically.
\end{itemize}

OUTPUT:\\  
Produce an enhanced version of the summary. Consider the score and reason to make it, it is important that this version is different from the previous one.\\
Extremely important:
\begin{itemize}
    \item Return a plain text as output, no html
    \item Don't change any number, as I need them as they are in the original context.
\end{itemize}
Please don't modify the placeholders.

\end{tcolorbox}

\subsection{LLM ELO Evaluation of Insightfulness}
\label{sec:llm_eval_prompt}

This prompt has not been tuned using human data, as that would involve an unfair advantage to the model.

\begin{tcolorbox}[colback=gray!5!white, colframe=black, title=Prompt 9: ELO Insightfulness Comparison Prompt, breakable]

Compare and return what insight according to the following criteria:\\
Insights must provide information that is:
\begin{itemize}
    \item Insightful. The information must be interesting and relevant.
    \item Actionable and impactful. The information must lead to tangible follow up actions and measurable outcomes
    \item Data driven. The information must be concrete and derived from real data.
\end{itemize}
\#\#\#\#\#\\
For the output, only return the best insight (either 'Insight 1' or 'Insight 2'). No yapping\\
\#\#\#\#\#\\
All insights are related to the database:\\
\{tables\_description\}\\
\#\#\#\#\#\\
Insight 1: \{insight1\}\\
\#\#\#\#\#\\
Insight 2: \{insight2\}

\end{tcolorbox}

\section{Database Descriptions}
For our task we used database descriptions generated with LLMs to simulate the worst case scenario. The following are the generated descriptions:

\begin{tcolorbox}[colback=gray!5!white, colframe=black, title=Description 1: california\_schools, breakable]

This database consists of three main tables: schools, satscores,
    and frpm. Each table contains a variety of columns with specific data formats
    and descriptions, providing detailed information about schools, SAT scores, and
    free or reduced-price meal (FRPM) eligibility.

    **schools Table:**

    This table provides comprehensive details about schools, including their identification
    numbers, status, location, contact information, and characteristics. Key columns
    include:

    - CDSCode: A unique identifier for each school.

    - NCESDist and NCESSchool: Identification numbers that, when combined, form a
    unique ID for each school according to the National Center for Educational Statistics.

    - StatusType: Indicates whether a school is active, closed, merged, or pending.

    - County, District, and School: Names of the county, district, and school.

    - Street, City, Zip, and State: Address details of the school.

    - MailStreet, MailCity, MailZip, and MailState: Mailing address details, which
    may default to the physical address if not provided.

    - Phone and Website: Contact information for the school.

    - OpenDate and ClosedDate: The dates a school opened or closed.

    - Charter, CharterNum, and FundingType: Information on whether a school is a charter
    and its funding type.

    - DOC and SOC: Codes indicating the district and school ownership types.

    - EdOpsCode, EdOpsName, EILCode, and EILName: Codes and names describing the educational
    options and instruction levels offered.

    - GSoffered and GSserved: The grade spans offered and served by the school.

    - Virtual and Magnet: Indicators of whether the school offers virtual instruction
    or is a magnet school.

    - Latitude and Longitude: Geographic coordinates of the school.

    - Administrative information: Names and email addresses of administrators.

    **satscores Table:**

    This table focuses on SAT performance data for schools, including:

    - cds: A reference to the CDSCode in the schools table.

    - sname, dname, cname: The names of the school, district, and county.

    - enroll12: Enrollment numbers for 12th grade.

    - NumTstTakr: The number of students who took the SAT.

    - AvgScrRead, AvgScrMath, AvgScrWrite: Average scores in reading, math, and writing.

    - NumGE1500: The number of students scoring 1500 or above on the SAT.

    **frpm Table:**

    This table provides data on free or reduced-price meal eligibility, a key indicator
    of student poverty levels, including:

    - CDSCode: A unique identifier for each school, linking back to the schools
    table.

    - Academic Year, County Code, District Code, School Code: Identifiers and the
    academic year of the data.

    - County Name, District Name, School Name: Names of the county, district, and
    school.

    - District Type, School Type, Educational Option Type: Descriptions of the school
    and district types.

    - NSLP Provision Status, Charter School (Y/N), Charter School Number, Charter
    Funding Type: Information on meal program status and charter school details.

    - Enrollment (K-12) and Enrollment (Ages 5-17): Enrollment numbers.

    - Free Meal Count and FRPM Count: Numbers of students eligible for free meals
    or reduced-price meals, for both K-12 and ages 5-17 categories.

    Each table serves a specific purpose, from providing detailed profiles of schools
    to reporting on academic performance and socio-economic indicators. Together,
    these tables offer a comprehensive view of the educational landscape, including
    school characteristics, student performance on standardized tests, and measures
    of economic disadvantage.

\end{tcolorbox}

\begin{tcolorbox}[colback=gray!5!white, colframe=black, title=Description 2: codebase\_community, breakable]

This database consists of multiple tables that store various
    types of data related to posts, users, comments, votes, tags, and badges within
    a community or forum-like platform. Each table is structured with columns that
    have specific data formats and descriptions, providing a comprehensive overview
    of the interactions and content within the platform.

    1. **postLinks**: This table keeps track of links between posts. Each entry
    has a unique post link ID, creation date, post ID, related post ID, and link type
    ID. The link type ID specifies the nature of the link between the posts.

    2. **postHistory**: This table records the history of edits or changes made
    to posts. It includes information such as the post history ID, post history type
    ID, post ID, revision globally unique ID (GUID), creation date of the edit, user
    ID of the editor, detailed content of the post after the edit, comments on the
    edit, and the editor's display name.

    3. **badges**: This table lists the badges awarded to users. Each badge has
    a unique ID, and the table records the user ID of the recipient, the badge name,
    and the date it was awarded.

    4. **posts**: This table contains detailed information about each post, including
    post ID, post type ID, accepted answer ID, creation date, score, view count, body
    text, owner user ID, last activity date, title, tags, answer count, comment count,
    favorite count, last editor user ID, last edit date, community owned date, parent
    post ID, closed date, owner display name, and last editor display name. The table
    provides a comprehensive overview of the content and interaction metrics for each
    post.

    5. **users**: This table stores information about users, including user ID,
    reputation, account creation date, display name, last access date, website URL,
    location, self-introduction, views, upvotes, downvotes, account ID, age, and profile
    image URL. This table helps in understanding user engagement and influence within
    the platform.

    6. **tags**: This table details the tags used within the platform, including
    tag ID, tag name, count of posts containing the tag, excerpt post ID, and wiki
    post ID. The count indicates the popularity of each tag.

    7. **votes**: This table records votes cast by users on posts. It includes
    vote ID, post ID, vote type ID, creation date of the vote, user ID of the voter,
    and bounty amount (if any). This table is crucial for understanding user preferences
    and the quality of content.

    8. **comments**: This table captures comments made on posts. It includes comment
    ID, post ID, score, detailed content of the comment, creation date, and user ID
    and display name of the commenter. The score provides insight into the sentiment
    or quality of the comment.

    Each table is interconnected through various IDs, allowing for a relational structure
    that supports complex queries and analysis. This database structure enables the
    platform to store, retrieve, and analyze data related to user interactions, content
    quality, and community engagement effectively.

\end{tcolorbox}

\begin{tcolorbox}[colback=gray!5!white, colframe=black, title=Description 3: debit\_card\_specializing, breakable]

This database consists of several tables that store information related to customers, gas stations, products, consumption patterns over time, and transactions. Each table is structured with columns that have specific data formats and descriptions. Here's a breakdown of each table and the data it contains:
    
1. **Customers Table ('customers'):**
   - **CustomerID:** This is an integer field that uniquely identifies each customer.
   - **Segment (client segment):** A text field that categorizes customers into different segments.
   - **Currency:** This text field specifies the currency used by the customer.

2. **Gas Stations Table ('gasstations'):**
   - **GasStationID (Gas Station ID):** An integer field uniquely identifying each gas station.
   - **ChainID (Chain ID):** An integer field that identifies the chain to which the gas station belongs.
   - **Country:** A text field indicating the country where the gas station is located.
   - **Segment (chain segment):** A text field categorizing the gas station into different segments.

3. **Products Table ('products'):**
   - **ProductID (Product ID):** An integer field uniquely identifying each product.
   - **Description:** A text field providing a description of the product.

    4. **YearMonth Table ('yearmonth'):**
   - **CustomerID (Customer ID):** An integer field linking the record to a specific customer.
   - **Date:** A text field indicating the date.
   - **Consumption:** A real number field representing the consumption amount.

5. **Transactions Table ('transactions\_1k'):**
   - **TransactionID (Transaction ID):** An integer field uniquely identifying each transaction.
   - **Date:** A date field indicating when the transaction occurred.
   - **Time:** A text field specifying the time of the transaction.
   - **CustomerID (Customer ID):** An integer field linking the transaction to a specific customer.
   - **CardID (Card ID):** An integer field identifying the card used for the transaction.
   - **GasStationID (Gas Station ID):** An integer field linking the transaction to a specific gas station.
   - **ProductID (Product ID):** An integer field identifying the product involved in the transaction.
   - **Amount:** An integer field indicating the quantity of the product.
   - **Price:** A real number field indicating the price of the product. It's noted that the total price can be calculated as the product of the Amount and Price fields.

Each table serves a specific purpose, from identifying customers and gas stations to detailing transactions and product descriptions. The structure allows for comprehensive data analysis across different dimensions such as customer behavior, product sales, and consumption patterns over time.

\end{tcolorbox}

\begin{tcolorbox}[colback=gray!5!white, colframe=black, title=Description 4: european\_football\_2, breakable]

This database consists of several tables that store detailed
    information about football (soccer) teams, players, matches, leagues, and countries.
    Each table is structured with columns that have specific data formats and descriptions,
    providing a comprehensive dataset for analysis or application development in the
    context of football.

    \#\#\# Team Attributes ('Team\_Attributes')

    This table contains attributes related to football teams, including their FIFA
    API ID, team API ID, and various characteristics like build-up play speed, dribbling,
    passing, and defensive strategies. Attributes are categorized into numerical values
    and classes (e.g., Slow, Balanced, Fast for speed) to describe the team's playing
    style and strategies in different aspects of the game.

    \#\#\# Player ('Player')

    The Player table stores basic information about football players, such as their
    unique IDs (both API and FIFA), name, birthday, height, and weight. This table
    can be used to identify players and their physical attributes.

    \#\#\# Match ('Match')

    This table records details of football matches, including unique IDs for the match,
    country, and league, as well as the season, stage, and date of the match. It also
    includes the home and away team IDs, goals scored by each team, and detailed player
    positions and actions during the match (e.g., shots on goal, fouls committed).
    Betting odds from various betting agencies are also stored here, providing insights
    into the expected outcomes of the matches.

    \#\#\# League ('League')

    The League table contains a unique ID for each league, along with the country
    ID and the name of the league. This table can be used to identify different football
    leagues and their associated countries.

    \#\#\# Country ('Country')

    This table simply maps each country's unique ID to its name, allowing for easy
    identification of countries in the dataset.

    \#\#\# Player Attributes ('Player\_Attributes')

    Similar to the Team Attributes table but for individual players, this table includes
    detailed ratings and scores for various skills and attributes (e.g., crossing,
    finishing, dribbling, defensive skills) as well as the player's preferred foot,
    attacking and defensive work rates. These attributes are rated numerically, typically
    on a scale from 0 to 100, providing a quantitative measure of a player's abilities
    and potential.

    \#\#\# Team ('Team')

    The Team table stores information about football teams, including their unique
    IDs (both API and FIFA), as well as their long and short names. This table can
    be used to identify teams and their official names.

    Each table is designed to provide a detailed dataset that can be used for statistical
    analysis, game prediction, player evaluation, and other applications related to
    football. The structure and organization of the tables allow for complex queries
    and analyses, making it a valuable resource for researchers, analysts, and enthusiasts
    interested in the beautiful game.

\end{tcolorbox}

\begin{tcolorbox}[colback=gray!5!white, colframe=black, title=Description 5: student\_club, breakable]

This database is designed to manage and track the financial, operational, and membership activities of an organization, likely a club or a small non-profit entity. It integrates several key functions related to income, expenses, budget management, events, membership, and geographic data. By centralizing this information, the database facilitates effective financial planning, monitoring of member involvement, and coordination of events.

    Income Table
    
    Description: Records the details of income received by the organization, including the source, amount, and associated member.
    
    Columns:
    
    Income ID (income\_id): A unique identifier for each record of income (Text).
    Date Received (date\_received): The date the funds were received (Text).
    Amount (amount): The amount of funds received (Integer, unit: dollar).
    Source (source): Indicates where the funds came from, such as dues or university allocation (Text).
    Notes (notes): Free-text field for additional details about the income (Text).
    Link to Member (link\_to\_member): Links to the member associated with the income (Text).
    
    Budget Table
    
    Description: Contains budget information for various categories and events, tracking amounts budgeted, spent, and remaining.
    
    Columns:
    
    Budget ID (budget\_id): A unique identifier for each budget entry (Text).
    Category (category): The budgeted category, e.g., advertisement, food, parking (Text).
    Spent (spent): The total amount spent in the budgeted category (Real, unit: dollar).
    Remaining (remaining): Calculated as the amount budgeted minus the amount spent (Real, unit: dollar).
    Amount (amount): The total budgeted amount for the specified category and event (Integer, unit: dollar).
    Event Status (event\_status): The current status of the event (Closed/Open/Planning) (Text).
    Link to Event (link\_to\_event): Links to the associated event (Text).
    
    Zip\_Code Table
    
    Description: Stores ZIP code information, including the associated city, county, and state.
    
    Columns:
    
    Zip Code (zip\_code): The five-digit ZIP code (Integer).
    Type (type): The type of ZIP code (Standard/PO Box/Unique) (Text).
    City (city): The city associated with the ZIP code (Text).
    County (county): The county associated with the ZIP code (Text).
    State (state): The state associated with the ZIP code (Text).
    Short State (short\_state): The abbreviated state name (Text).
    
    Expense Table
    
    Description: Details expenses incurred by the organization, including amounts, descriptions, and approval status.
    
    Columns:
    
    Expense ID (expense\_id): A unique identifier for each expense record (Text).
    Expense Description (expense\_description): A description of what the money was spent on (Text).
    Expense Date (expense\_date): The date the expense was incurred (Text, format: YYYY-MM-DD).
    Cost (cost): The dollar amount of the expense (Real, unit: dollar).
    Approved (approved): Indicates whether the expense was approved (True/False) (Text).
    Link to Member (link\_to\_member): Links to the member associated with the expense (Text).
    Link to Budget (link\_to\_budget): Links to the related budget entry (Text).
    
    Member Table
    
    Description: Contains information about members, including personal details, position, and contact information.
    
    Columns:
    
    Member ID (member\_id): A unique identifier for each member (Text).
    First Name (first\_name): The member's first name (Text).
    Last Name (last\_name): The member's last name (Text).
    Email (email): The member's email address (Text).
    Position (position): The position held by the member in the organization (Text).
    T-Shirt Size (t\_shirt\_size): The member's preferred t-shirt size (Text).
    Phone (phone): The member's contact phone number (Text).
    Zip (zip): The ZIP code of the member's hometown (Integer).
    Link to Major (link\_to\_major): Links to the member's major (Text).
    
    Attendance Table
    
    Description: Tracks attendance of members at various events.
    
    Columns:
    
    Link to Event (link\_to\_event): Links to the event attended (Text).
    Link to Member (link\_to\_member): Links to the member who attended the event (Text).
    
    Event Table
    
    Description: Contains details about events organized by the organization, including dates, types, and locations.
    
    Columns:
    
    Event ID (event\_id): A unique identifier for each event (Text).
    Event Name (event\_name): The name of the event (Text).
    Event Date (event\_date): The date the event took place or is scheduled to take place (Text, format: YYYY-MM-DD).
    Type (type): The type of event (e.g., game, social, election) (Text).
    Notes (notes): Additional notes about the event (Text).
    Location (location): The address or name of the event location (Text).
    Status (status): The current status of the event (Open/Closed/Planning) (Text).
    
    Major Table
    
    Description: Stores information about academic majors, including the associated department and college.
    
    Columns:
    
    Major ID (major\_id): A unique identifier for each major (Text).
    Major Name (major\_name): The name of the major (Text).
    Department (department): The name of the department offering the major (Text).
    College (college): The name of the college housing the department (Text).

\end{tcolorbox}

\section{Cost Estimation}
\label{sec:cost}
Using GPT4o, the SQLAgent has a mean price of \$0.072 for each query.
The Hypothesis Generator for our method has a mean price of \$0.113 per use.

On average, each insight generated by our method has 7.25 subquestions, resulting in a mean cost of \$0.637 per insight.

This is a great improvement over the previous manual method of generating insights, which amounts to an approximate of \$140 per insight (as estimated in our company).

Additionally, the cost of GPT-4o has recently dropped by 79\% in a few months. We therefore expect the cost of LLM to continue to drop, given also recent models (e.g., Deepseek) that can reach top performance at a much lower cost — 10x to 30x lower compared to GPT-4o.

\section{Use of Coding Assistant}
ChatGPT was used as a coding assistant for the elaboration of the results and the experiments.

\end{document}